\documentclass[preprint,12pt]{elsarticle}

\usepackage{amssymb}
\usepackage{amsmath}
\usepackage{url}
\usepackage{graphicx}
\usepackage{bbding}
\usepackage{bbm}
\usepackage{multirow}
\usepackage{booktabs}
\usepackage{adjustbox}
\usepackage{setspace}

\journal{Brain Hemorrhages}

\begin{document}
	
\begin{frontmatter}

\title{Zero-Shot Multi-modal Large Language Models v.s. Supervised Deep Learning: A Comparative Analysis on CT-Based Intracranial Hemorrhage Subtyping}

\renewcommand{\thefootnote}{}

\author[1]{Yinuo Wang, PhD} 
\author[2]{Kai Chen, PhD}
\author[1]{Yue Zeng, MSc}
\author[1]{Cai Meng, PhD}
\author[3]{Chao Pan, PhD}
\author[3]{Zhouping Tang, PhD}\footnotetext{Corresponding author: Cai Meng, Tsai@buaa.edu.cn}
\affiliation[1]{organization={Image Processing Center, Beihang University},
	city={Beijing 100191},
	country={China}}
\affiliation[2]{organization={School of Mechanical Engineering and Automation, Beihang University},
	city={Beijing 100191},
	country={China}}
\affiliation[3]{organization={Department of Neurology, Tongji Hospital, Tongji Medical College, Huazhong University of Science and Technology},
	city={Wuhan 430030},
	country={China}}

\begin{abstract}
	\textbf{Background and Purpose:} Accurate identification of intracranial hemorrhage (ICH) subtypes on non-contrast CT is crucial for prognosis and treatment but remains challenging due to low contrast and blurred boundaries. This study evaluates the zero-shot performance of multi-modal large language models (MLLMs) versus traditional deep learning in ICH detection and subtyping.
	\textbf{Materials and Methods:} Using 192 NCCT volumes from the RSNA dataset, we compared MLLMs (GPT-4o, Gemini 2.0 Flash, Claude 3.5 Sonnet V2) with deep learning models (ResNet50, Vision Transformer). MLLMs were prompted for ICH presence, subtype, localization, and volume estimation.
	\textbf{Results:} Traditional deep learning models outperformed MLLMs in both ICH detection and subtyping. For subtyping, MLLMs showed lower accuracy, with Gemini 2.0 Flash achieving a macro-averaged precision of 0.41 and F1 score of 0.31.
	\textbf{Conclusions:} While MLLMs offer enhanced interpretability through language-based interaction, their accuracy in ICH subtyping remains inferior to deep learning networks. Further optimization is needed to improve their utility in three-dimensional medical imaging.
\end{abstract}



\begin{highlights}
	\item Multi-modal large language models (MLLMs) show potential in intracranial hemorrhage (ICH) diagnosis and treatment with visual and interactive capabilities.
	\item Proprietary and open-source MLLMs still underperform in ICH binary/subtype classification compared to trained deep networks.
	\item MLLMs require fine-tuning for improved ICH subtyping and related tasks.
\end{highlights}


\end{frontmatter}

\section{Introduction}
\label{introduction}

Intracranial hemorrhage (ICH) is a life-threatening acute cerebrovascular disease, affecting approximately 2 million individuals worldwide each year. Despite advances in medical care, the 30-day mortality rate remains exceptionally high, ranging between 35\% and 52\%, with only one-fifth of survivors achieving complete recovery within six months post-onset \cite{Incidence, AdvanceinICH}. The etiology of ICH is multifactorial, commonly associated with external trauma, hypertension, thrombosis, neoplasms, or vascular malformations \cite{acutemanagement,BH1-SAH, BH2-Spontaneous}. These underlying causes can lead to one or more subtypes of hemorrhage, which are typically classified by location into epidural hemorrhage (EDH), intraparenchymal hemorrhage (IPH), intraventricular hemorrhage (IVH), subarachnoid hemorrhage (SAH), and subdural hemorrhage (SDH) \cite{riskfactors, cSDH}.
During the acute phase of ICH, non-contrast computed tomography (NCCT) scans are widely used to detect and characterize specific hemorrhage subtypes, facilitating the development of timely and personalized treatment strategies. Moreover, NCCT imaging remains essential during surgical procedures and follow-up evaluations to assess residual hemorrhage and monitor the recovery process \cite{neuroimaginginICH, IndiForMIC, BH3-neuroimaging}.
Therefore, the rapid and accurate identification of ICH and its subtypes based on NCCT is critical for hospitals worldwide. This is particularly challenging in resource-limited settings, where access to highly experienced clinicians and advanced imaging technologies is often restricted.

The detection of ICH subtypes is typically complicated by the variability in the shape, size, and contrast with surrounding soft tissues. This variability increases the risk of missed or incorrect diagnoses during NCCT evaluations. Furthermore, studies have demonstrated notable inter-observer variability among radiologists with varying levels of expertise, which negatively impacts patient management and clinical outcomes \cite{Misidentification}.
In recent years, advancements in artificial intelligence have driven the development of deep learning methods for ICH subtype diagnosis. Early approaches utilized convolutional neural networks (CNNs) to extract low-level features from slices \cite{ExplainableICH, 3dsubtypeclassi, EnsembledDL, SixClassification, cnn-rnn-attention, evaluation-cnnrnn}.
Lee et al. emulated the diagnostic workflow of radiologists by implementing multi-window conversion and slice interpolation as preprocessing steps for NCCT scans. They incorporated pretrained CNNs for ICH subtype prediction, complemented by attention maps to enhance the interpretability of predictions \cite{ExplainableICH}.
Ye et al. developed a joint CNN-RNN classification framework, which used a CNN to extract slice-level features and an RNN to model contextual relationships across slices. The framework first performed binary ICH detection, followed by five-class subtype classification on positive samples to identify the specific ICH subtypes present \cite{3dsubtypeclassi}.
Subsequent research has advanced classification methodologies through multiple directions. Notably, Vision Transformers (ViTs) have been increasingly adopted as backbone architectures to replace conventional CNNs, demonstrating enhanced non-local feature representation capabilities in medical imaging tasks \cite{vitfusion}.
To address computational constraints, lightweight modules have been developed, achieving comparable accuracy with reduced parameters \cite{EfficientNN}. 
The integration of attention mechanisms has further improved model performance by selectively focusing on pathologically significant regions \cite{cnn-rnn-attention, GELTTA}.
Technical refinements in preprocessing pipelines, particularly artifact reduction strategies, have enhanced ICH detection accuracy through minimized noise interference \cite{artifactreduction}. 
Meanwhile, semi-supervised learning frameworks leveraging unlabeled datasets effectively mitigate data scarcity challenges while maintaining diagnostic robustness \cite{Semi-Supervised}.
These synergistic advancements collectively contribute to more reliable and deployable computer-aided ICH diagnosis systems.

Despite significant advancements in classification and detection accuracy, these primary automated methods predominantly generates elementary outputs of predicted classifications or detection results, failing to interact with human like a reliable expert.
That means, when an ICH subtype is recognized by these methods, radiologists must still meticulously analyze and assess bleeding volume, severity, treatment options, and other abnormalities, which continues to demand clinical expertise and labor-intensive scrutiny from medical personnel.
With the advent of artificial general intelligence (AGI), multi-modal large language models (MLLMs) are poised to revolutionize radiology workflows. These systems synergize robust visual processing with natural language capabilities that encapsulate accumulated diverse medical knowledge, thereby enabling intuitive interaction between generalized AI models and end users \cite{MMMUBenchmark, GMAI-VL}.
Across diverse medical applications, MLLMs demonstrate the capacity to provide valuable differential diagnoses from input images, generate preliminary structured reports, characterize lesions, evaluate disease burden, and recommend appropriate treatment and follow-up protocols \cite{Med-MLLM, Generalist, NaturePathology, NatureOncology, EndoChat}. These capabilities can assist radiologists in triage, prioritization, and decision-support processes. Within this context, our research aims to evaluate the efficacy of contemporary MLLMs in identifying ICH subtypes.
In this study, we emulate the diagnostic process of neurosurgeons by inputting windowed NCCT slices alongside progressive prompts to guide MLLMs in predicting ICH presence, subtype classification, localization, hemorrhage volume, treatment recommendations, and other potential abnormalities. A comprehensively annotated dataset comprising 192 NCCT scans with 6,404 slices was partitioned into training, validation, and testing subsets. Our evaluation on the test set encompasses not only proprietary MLLMs and more lightweight, deployable open-source alternatives but also incorporates comparative analysis with conventional deep learning classification models based on CNNs and transformer architectures.

\section{Methods}
\subsection{Data Usage}

\begin{figure}[ht]
\centering
\adjustbox{center, max width=0.5\textwidth}{\includegraphics{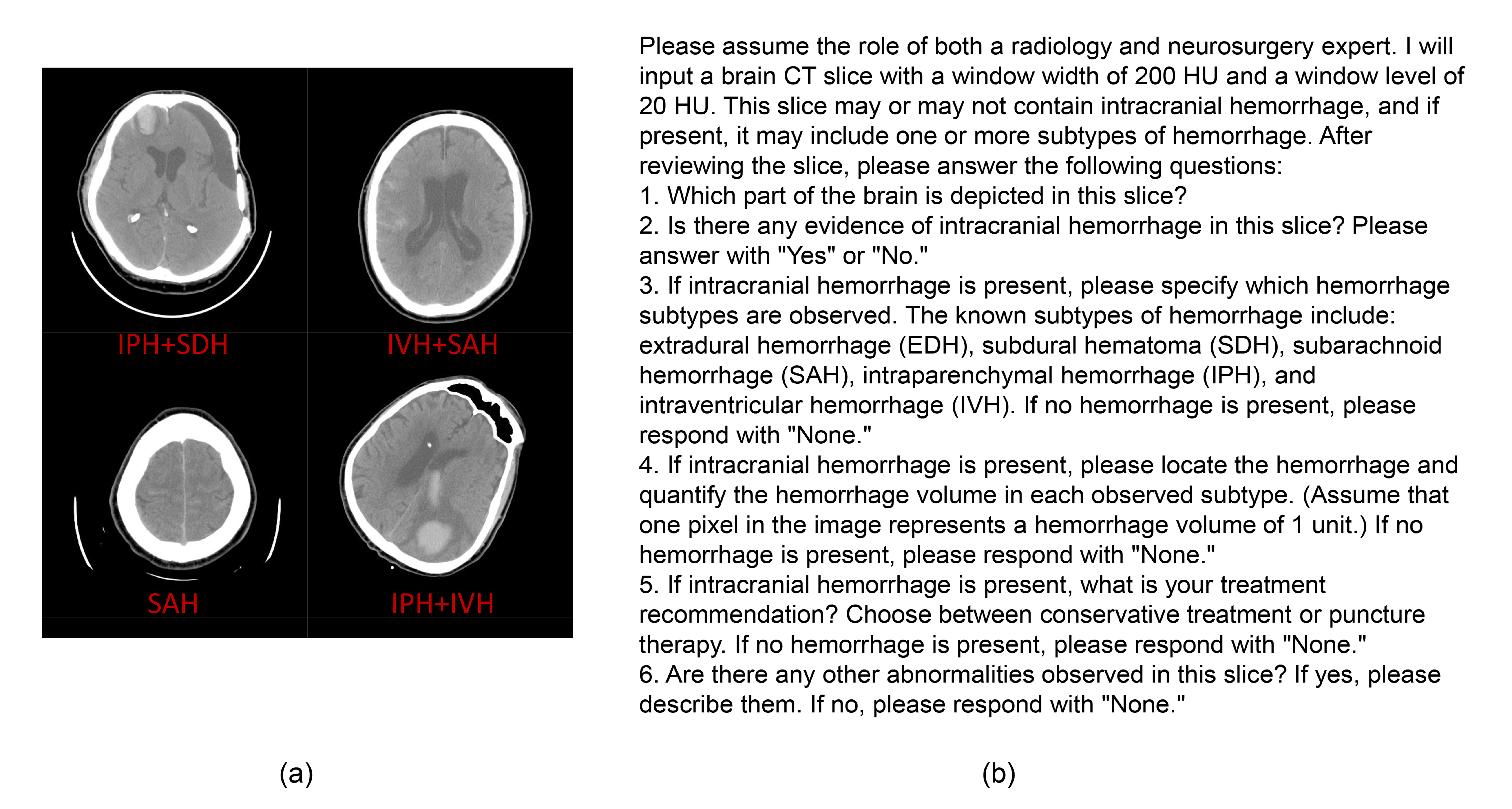}}
\caption{(a) Random samples from the utilized dataset, with hemorrhage subtypes labeled in red; (b) The designed prompt.}
\label{fig0}
\end{figure}

This study was approved by the Biomedical Ethics Committee of Beihang University (Approval Number: BM20240182).
We employed a subset of the brain hemorrhage dataset originally provided by the Radiological Society of North America (RSNA). The dataset was initially released as randomly ordered DICOM-format slices. Wu et al. \cite{BHSD} reconstructed 192 NCCT volumes, each containing one or more subtypes of ICH, by leveraging anonymized patient identifiers and geometric/positional metadata embedded in the DICOM files. These volumes were subsequently annotated at the pixel level by two medical imaging experts and radiology residents. The annotation process was guided by the original slice-level hemorrhage annotations and included five ICH subtypes: EDH, IPH, IVH, SAH, and SDH. The dataset was made available under the MIT License.
During preprocessing, windowing was applied to the volumes with a window width of 200 and a window level of 20. Slices above the cranial vertex that lacked intracranial tissue were excluded, and the remaining slices were converted to PNG format. For experimental purposes, 10\% of the slices were randomly selected to create the test set, resulting in a total of 561 slices.

\subsection{Prompt Usage}

Prompts act as a pivotal mechanism connecting user intent with model capabilities in large language models (LLMs), directing the generation of desired outputs through carefully crafted contextual cues \cite{Prompt1, Prompt2}. In multi-modal applications, prompts leverage cross-modal information, such as text and images, to expand the model's ability to generalize across complex tasks. Moreover, task-specific prompt designs can substantially enhance output quality by achieving an optimal balance between factual accuracy, creativity, and controllability.

This study introduces a progressive prompt design for the cases, as depicted in Fig.~\ref{fig0} b. Building on prior prompting methodologies (citation), we initially assigned specific roles to the MLLMs to activate their domain-specific knowledge, instructing them to "act as a radiology expert and a neurosurgery specialist." Following the introduction of task-relevant context, the first question focused on identifying the brain region represented in the given slice, aimed at evaluating the MLLMs' scene comprehension capabilities. The second question examined the presence of intracranial hemorrhage within the slice, framing the task as a binary classification problem. The third question expanded upon the second, requiring the MLLMs to identify the subtype of intracranial hemorrhage if detected. The fourth question further progressed by instructing the MLLMs to localize the hemorrhage area and quantify its volume.
The final two questions shifted toward clinical decision-making and anomaly detection. Specifically, they assessed the MLLMs' preferences for treatment recommendations and their ability to recognize potential disease threats. These conclusions, designed as open-ended interactions, lacked ground truth comparisons, allowing for exploratory evaluation. This progressive prompt design aimed to systematically assess the MLLMs' capabilities across binary classification, multi-class classification, localization, and quantification tasks related to intracranial hemorrhages. Furthermore, the open-ended interactions were intended to explore the MLLMs' reasoning concerning the intrinsic links between diagnostic processes and clinical decision-making.

\subsection{Multi-Modal Large Language Models (MLLMs)}
This study employed both proprietary and open-source MLLMs. The proprietary models include GPT-4o, Gemini 2.0 Flash, and Claude 3.5 Sonnet V2, while the open-source models comprise Qwen-VL-3b-Instruct, DeepSeek-VL2-Tiny, and LLaVA-Med-v1.5-Mistral-7b. GPT-4o was developed by OpenAI and released in May 2024, Gemini 2.0 Flash by Google in January 2025, and Claude 3.5 Sonnet V2 by Anthropic in October 2024. For the open-source models, Qwen-VL-3b-Instruct was launched by Alibaba Cloud in February 2025, DeepSeek-VL2-Tiny by DeepSeek in December 2024, and LLaVA-Med-v1.5-Mistral-7b by Microsoft in May 2024. Notably, LLaVA-Med-v1.5-Mistral-7b is specifically tailored for applications in the biomedicine domain, whereas the other models are designed for general-domain tasks.

During the evaluation phase, GPT-4o, Gemini 2.0 Flash, and Claude 3.5 Sonnet V2 were accessed through their respective application programming interfaces (APIs), while Qwen-VL-3b-Instruct, DeepSeek-VL2-Tiny, and LLaVA-Med-v1.5-Mistral-7b were deployed locally on a server equipped with four Nvidia RTX 3090 GPUs. To ensure consistent and deterministic outputs, the temperature parameter for all models was set to 0.1.

\subsection{Learning-Based Classifiers}
To comprehensively assess the classification performance of MLLMs, we incorporated commonly used image classification models based on CNNs and Transformers, including ResNet50, ViT-B, ViT-L, and SwinTransformer-v2-B. For intracranial hemorrhage classification tasks, encompassing binary classification and subtype recognition, all models were trained under a unified strategy. Specifically, slices excluded from the test set were partitioned into training and validation sets in an 8:2 ratio. The training process was performed for 100 epochs, employing the cross-entropy loss function, Adam optimizer, a learning rate of $10^{-4}$, and a batch size of 32. To enhance model generalization, data augmentation techniques such as random flipping and rotation were applied. For the subtype recognition task, we utilized the multi-label binary cross-entropy loss function (log-loss), which is defined as follows:
\begin{equation}
\mathcal{L}_{} = -\frac{1}{K} \frac{1}{C} \sum_{k=1}^{K} \sum_{c=1}^{C} \left[ y_{c,k} \log(\hat{y}_{c,k}) + (1 - y_{c,k}) \log(1 - \hat{y}_{c,k}) \right]
\end{equation}
where $K$ and $C$ denotes the number of training samples and classes, respectively. $C$ equals to $5$ in our work, including five ICH subtypes. $y_{c,k}$ represents the ground truth label of the $k$-th sample for the $c$-th class, and $\hat{y}_{c,k}$ represents the predicted ICH subtype probabilities.

\section{Results}
\subsection{Performance of ICH Binary Classification}

\begin{table}[ht]
\centering
\caption{Quantitative comparison of ICH binary classification on the testing datasets. \textbf{Bold} text denotes the best performance in each group of methods.}
\label{tab:binaryclassification}
\resizebox{\textwidth}{!}{
	\begin{tabular}{@{}l|ccccc@{}}
		\toprule
		Methods & Accuracy & Precision & Sensitivity & Specificity & F1 score \\ \midrule
		\multicolumn{5}{c}{Proprietary}  \\ \midrule
		GPT-4o & 0.7005 & 0.6390 & {\bfseries0.6553} & 0.7331 & {\bfseries0.6471}  \\
		Gemini 2.0 Flash  & {\bfseries0.7504} & {\bfseries0.7987} & 0.5404 & {\bfseries0.9018} & 0.6447  \\
		Claude 3.5 Sonnet V2  & 0.6275 & 0.5489 & 0.6213 & 0.6319 & 0.5828 \\ \midrule
		\multicolumn{5}{c}{Open-source}  \\ \midrule
		Qwen-VL-3b-Instruct  & 0.4385 & 0.4231 & 0.9362 & 0.0798 & 0.5828   \\
		DeepSeek-VL2-Tiny  & 0.4225 & 0.4204 & {\bfseries1.0000} & 0.0061 & {\bfseries0.5919}  \\
		LLaVA-Med-v1.5-Mistral-7b  & {\bfseries0.4955} & {\bfseries0.4358} & 0.6936 & {\bfseries0.3528} & 0.5353  \\ \midrule
		\multicolumn{5}{c}{Classifiers}  \\ \midrule
		ResNet50  & 0.8966  & 0.8734 & {\bfseries0.8809} & 0.9080 & 0.8771  \\
		ViT-B  & 0.8556  & 0.8702 & 0.7702 & 0.9171 & 0.8172  \\
		ViT-L  & 0.8770  & 0.8517 & 0.8553 & 0.8926 & 0.8535  \\
		SwinTransformer-v2-B  & {\bfseries0.9091}  & {\bfseries0.9035} & 0.8766 & {\bfseries0.9325} & {\bfseries0.8898}  \\
		\bottomrule
\end{tabular}}
\end{table}

Initially, we assessed the performance of models on the ICH binary classification task. As depicted in Table~\ref{tab:binaryclassification}, among proprietary MLLMs, GPT-4o and Gemini 2.0 Flash generally outperformed Claude 3.5 Sonnet V2. Specifically, Gemini 2.0 Flash exceeded GPT-4o in Accuracy, Precision, and Specificity approximately by 0.05, 0.16, and 0.17, respectively. In contrast, GPT-4o demonstrated superior Sensitivity and F1 score, leading by 0.11 and 0.002, respectively. These results suggest that Gemini 2.0 Flash is more adept at identifying healthy cases and reducing false positives, whereas GPT-4o offers better coverage for hemorrhage cases, albeit with increased false alarms.
Conversely, the included open-source MLLMs exhibited inferior overall performance compared to proprietary MLLMs. Despite achieving higher sensitivity rates, their peak accuracy, precision, specificity, and F1 scores were substantially lower. Notably, models such as Qwen-VL-3B-Instruct and DeepSeek-VL2-tiny showed low Specificity and high Sensitivity, resulting in a high rate of false positive predictions and indicating their limited ability to effectively distinguish hemorrhagic features.
Furthermore, classifiers trained on ICH datasets consistently outperformed MLLMs across all evaluated metrics except Sensitivity. This indicates that while MLLMs have advanced interactive capabilities, their visual classification performance is still lacking compared to deep networks trained on specialized data, primarily due to insufficient generalization in hemorrhage detection tasks.

\subsection{Performance of ICH Subtype Classification}

\begin{table}[ht]
\centering
\caption{Quantitative comparison of ICH subtype classification on the testing datasets. The results for the five subtypes are calculated using macro-averaging. \textbf{Bold} text denotes the best performance in each group of methods.}
\label{tab:multiclassification}
\resizebox{\textwidth}{!}{
	\begin{tabular}{@{}l|cccc@{}}
		\toprule
		Methods  & Precision & Sensitivity & Specificity & F1 score \\ \midrule
		\multicolumn{4}{c}{Proprietary} \\ \midrule
		GPT-4o & 0.3243 & 0.2734 & 0.9232 & 0.2842  \\
		Gemini 2.0 Flash  & {\bfseries0.4093} & {\bfseries0.2909} & {\bfseries0.9526} & {\bfseries0.3116}  \\
		Claude 3.5 Sonnet V2  & 0.2459 & 0.2169 & 0.9091 & 0.2282 \\ \midrule
		\multicolumn{4}{c}{Open-source} \\ \midrule
		Qwen-VL-3b-Instruct  & {\bfseries0.1069} & 0.3393 & 0.6442 & 0.1524   \\
		DeepSeek-VL2-Tiny  & 0.0997 & {\bfseries0.6366} & 0.3281 & {\bfseries0.1547}  \\
		LLaVA-Med-v1.5-Mistral-7b  & 0.1035 & 0.2682 & {\bfseries0.7438} & 0.1393 \\ \midrule
		\multicolumn{4}{c}{Classifiers} \\ \midrule
		ResNet50  & 0.8488 & {\bfseries0.8542} & 0.9765 & 0.8512  \\
		ViT-B  & 0.7964 & 0.8065 & 0.9705 & 0.8001  \\
		ViT-L  & 0.8056 & 0.8113 & 0.9730 & 0.8071  \\
		SwinTransformer-v2-B  & {\bfseries0.8782} & 0.8384 & {\bfseries0.9799} & {\bfseries0.8569}  \\
		\bottomrule
\end{tabular}}
\end{table}

\begin{figure}[ht]
\centering
\adjustbox{center, max width=0.5\textwidth}{\includegraphics{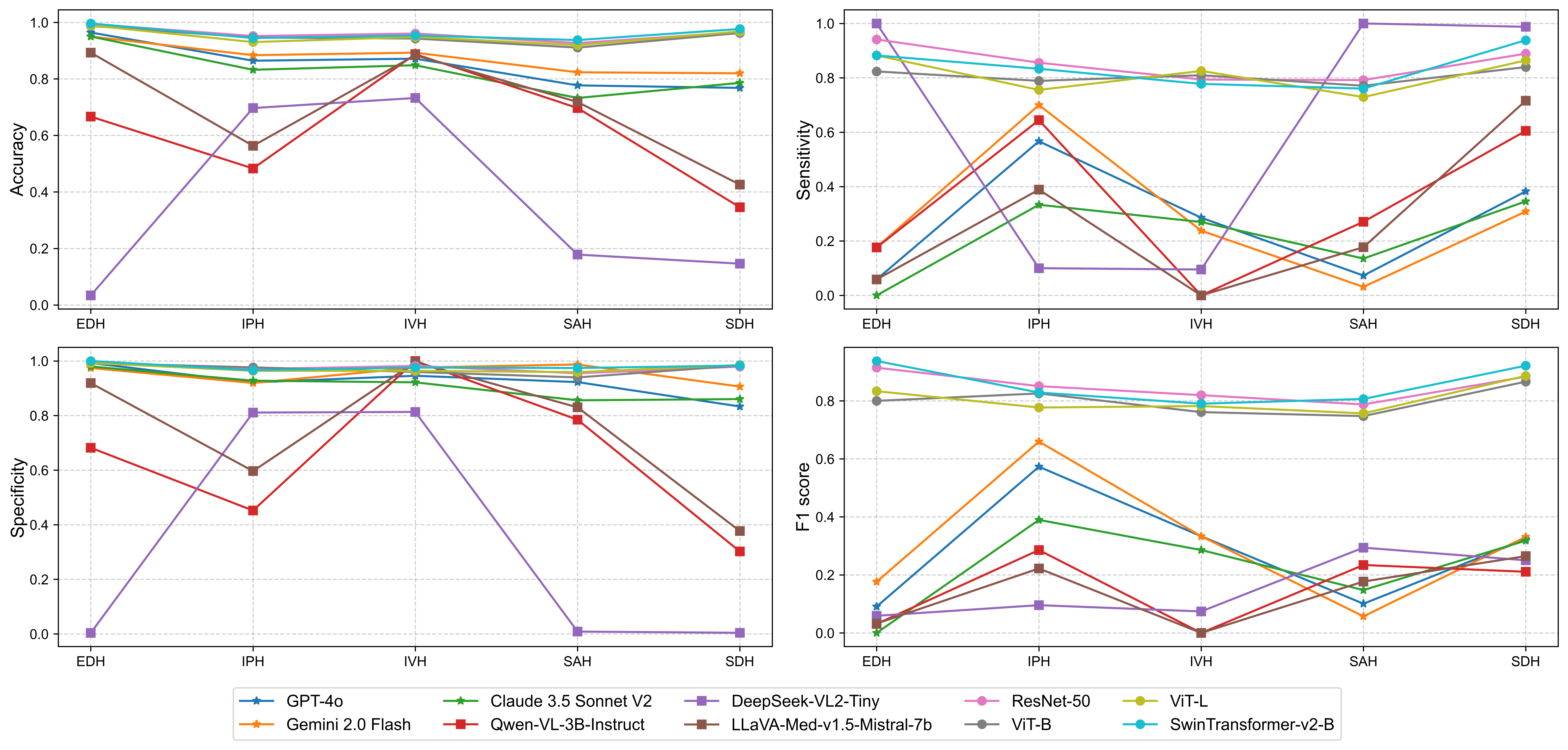}}
\caption{Line chart showing the performance of various models in the recognition of ICH subtypes.}
\label{fig1}
\end{figure}

Table~\ref{tab:multiclassification} presents the macro-average results of ten methods in classifying five intracranial hemorrhage (ICH) subtypes. In this task, MLLMs consistently underperform compared to learning-based classifiers. Among proprietary MLLMs, Gemini 2.0 Flash demonstrates the highest overall performance. However, it exhibits reductions of approximately 0.47 in Precision, 0.55 in Sensitivity, 0.03 in Specificity, and 0.55 in F1 score when compared to SwinTransformer-v2-B. This suggests that while proprietary MLLMs tend to generate fewer false positives, their proficiency in subtype identification is still limited. Notably in these comparisons, proprietary MLLMs generally surpass open-source MLLMs in performance.
Fig.~\ref{fig1} provides an independent analysis of ten methods for identifying the five subtypes. In terms of Sensitivity, MLLMs exhibit a significant decline compared to learning-based classifiers for EDH, IVH, and SAH, while the disparity is less pronounced for IPH and SDH. Regarding Specificity, proprietary MLLMs and learning-based classifiers show comparable performance across all five subtypes, aligning with observations in Table~\ref{tab:multiclassification}. For the F1 score, proprietary MLLMs excel in IPH but perform less effectively for EDH and SAH.

\subsection{Open-ended Analysis}

\begin{figure}[ht]
\centering
\adjustbox{center, max width=0.4\textwidth}{\includegraphics{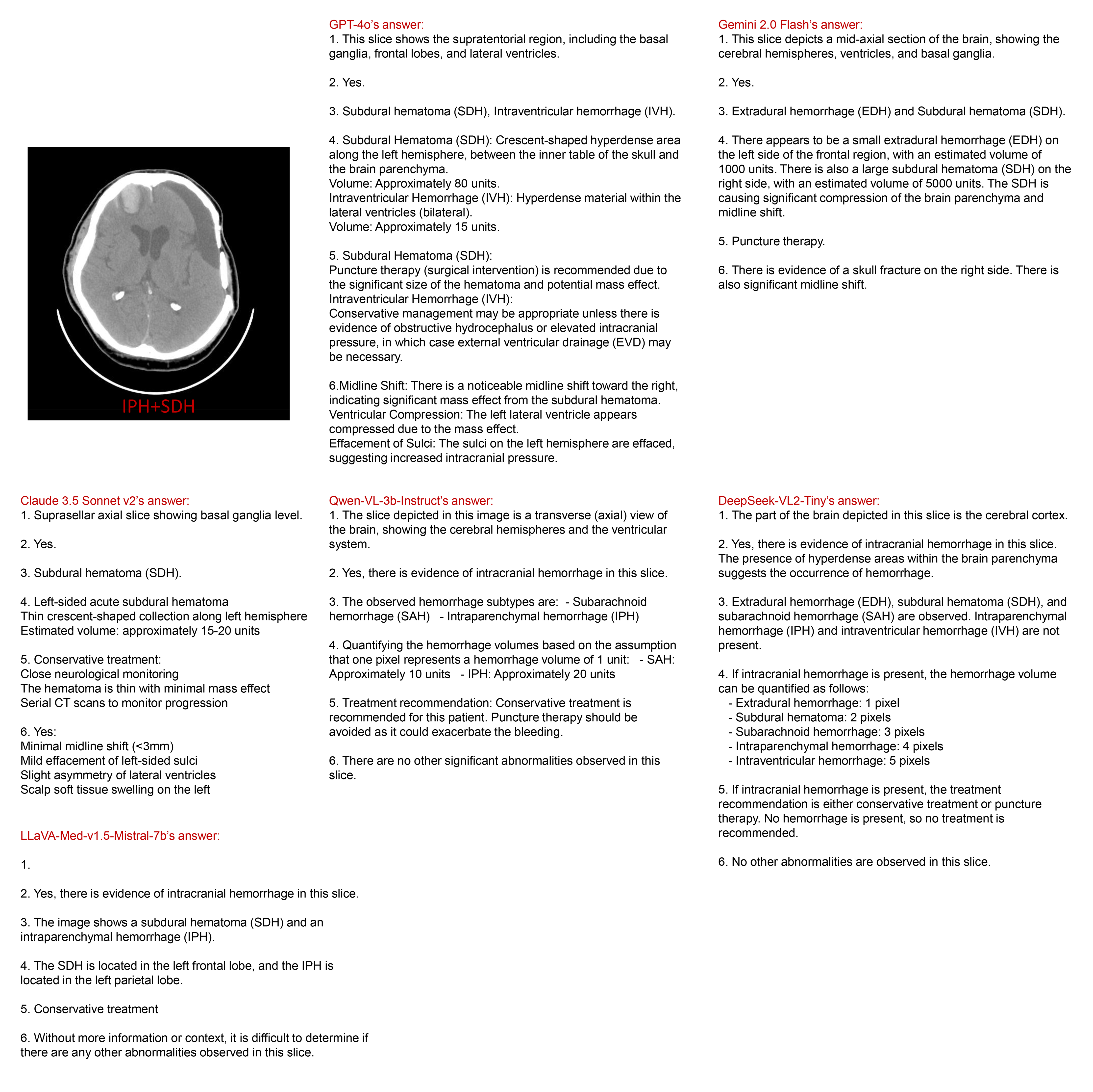}}
\caption{Comparison of responses from six MLLMs on a random case, which illustrates a fracture in the patient's left skull due to trauma, with SDH evident in the left hemisphere and IPH in the right hemisphere.}
\label{fig2}
\end{figure}

This section compares the responses of six MLLMs on specific cases to assess the comprehensiveness and reasoning of their answers. Fig.~\ref{fig2} illustrates an IPH in the right hemisphere and an SDH in the left hemisphere on the input slice. For this case, all MLLMs successfully detected the presence of hemorrhage (the second answer), but only LLaVA-Med-v1.5-Mistral-7b accurately identified the ICH subtypes (the third answer). Concerning the fourth question about hemorrhage volume, the ground truth annotations for IPH and SDH are 1600 and 5761 units, respectively. Only Gemini 2.0 Flash provided estimates within the correct order of magnitude but mistakenly identified IPH as EDH. For treatment recommendations in the fifth question, the models proposed various options: Gemini 2.0 Flash suggested puncture therapy, while Claude 3.5 Sonnet v2, Qwen-VL-3b-Instruct, and LLaVA-Med-v1.5-Mistral-7b recommended conservative treatment. GPT-4o advised conservative treatment for IVH and puncture therapy for SDH. Additionally, GPT-4o, Gemini 2.0 Flash, and Claude 3.5 Sonnet v2 reported observed abnormalities, such as midline shift, whereas the other three models reported no abnormalities. Overall, GPT-4o, Gemini 2.0 Flash, and Claude 3.5 Sonnet v2 provided more comprehensive and reasonably coherent responses to open-ended questions, while DeepSeek-VL2-Tiny's responses to questions three through five lacked coherence.

\section{Discussions}

\begin{figure}[ht]
\centering
\adjustbox{center, max width=0.4\textwidth}{\includegraphics{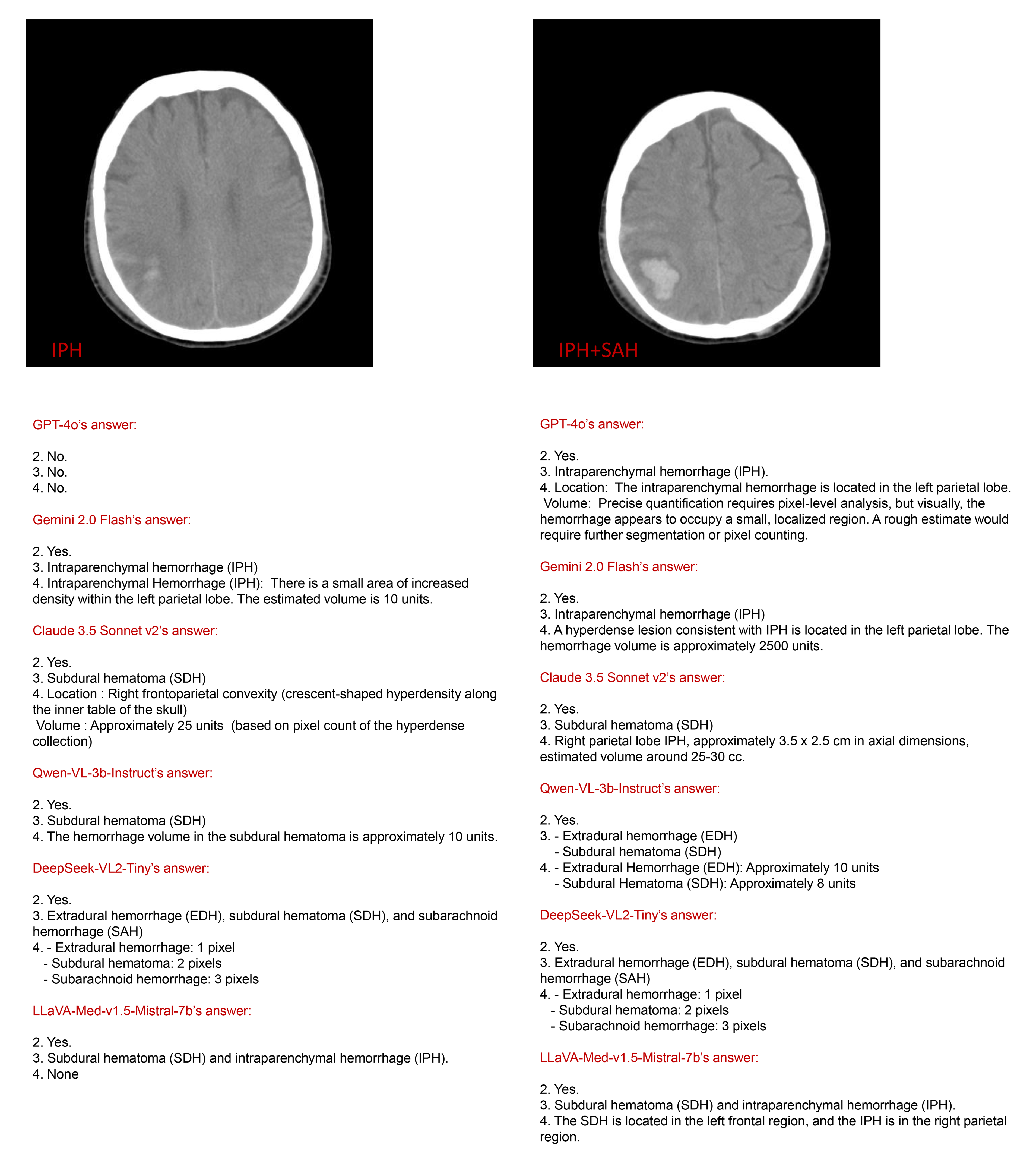}}
\caption{Comparison of responses from six MLLMs to two slices from different layers of the same NCCT volume.}
\label{fig3}
\end{figure}

\begin{figure}[ht]
\centering
\adjustbox{center, max width=0.4\textwidth}{\includegraphics{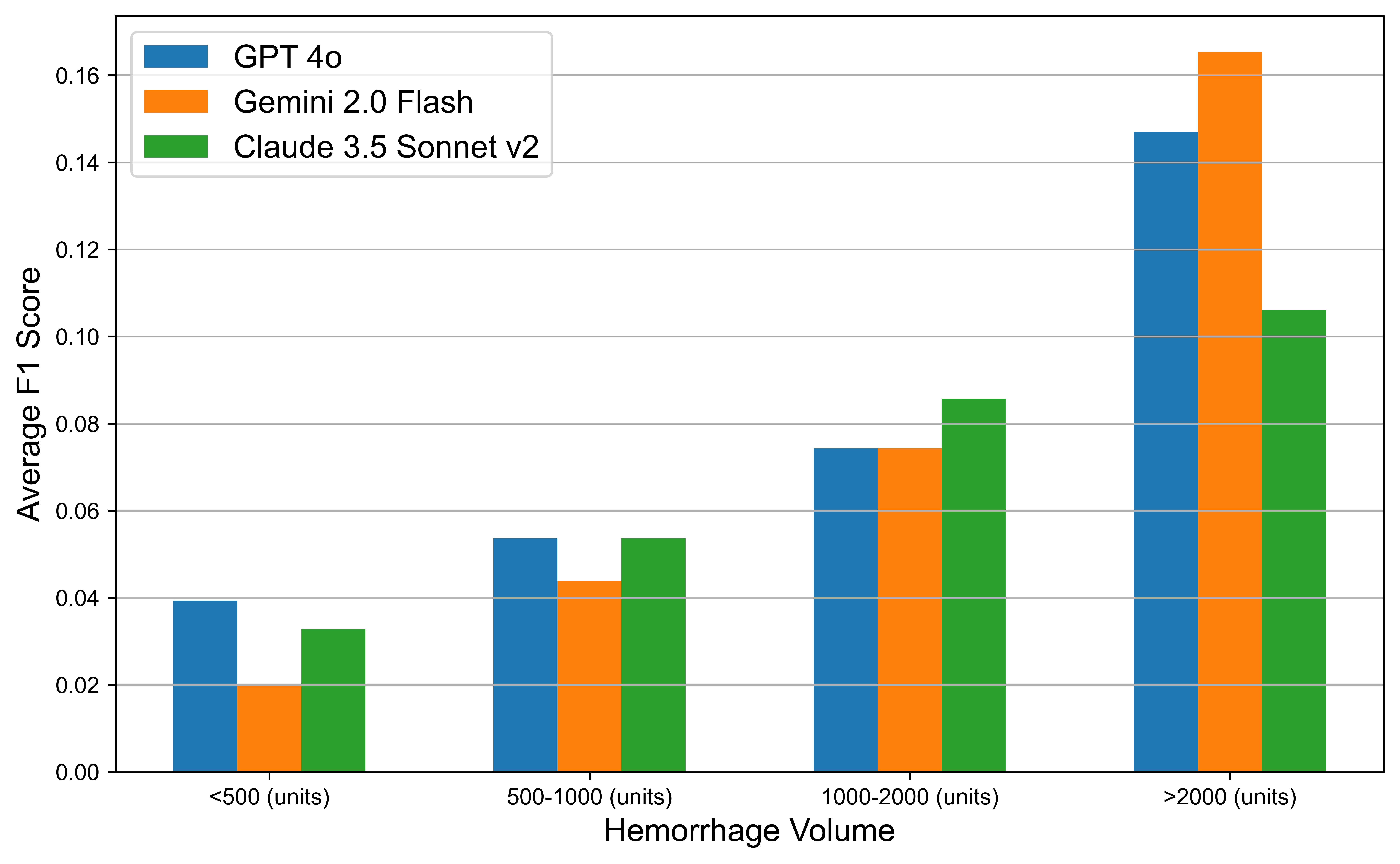}}
\caption{Statistical classification results of ICH subtypes across different hemorrhage volume ranges. The bar chart presents data for all positive cases, with the macro-averaged F1 score used for evaluation.}
\label{fig4}
\end{figure}

This prehospital study assessed the effectiveness of various MLLMs in identifying ICH subtypes. The results reveal that mainstream MLLMs, both proprietary and open-source, fall short in classification performance compared to trained classifiers. Proprietary MLLMs notably outperform open-source ones, likely due to the smaller parameter size of deployable MLLMs utilized in this study. 
Comparative analysis of Table~\ref{tab:binaryclassification} and Table~\ref{tab:multiclassification} demonstrates that MLLMs are significantly less effective in recognizing ICH subtypes than in detecting  general hemorrhages. This indicates that while MLLMs possess basic hemorrhage detection capabilities, they lack the ability to discern and learn specific subtype characteristics. Our observations suggest that the accuracy of MLLMs in identifying ICH correlates with the contrast and size of the lesion, consistent with intuitive principles of image processing.
Fig.~\ref{fig3} illustrates two cases from different slices of the same NCCT volume, where the contrast of IPH lesion in the left image is lower than in the right. Consequently, GPT-4o failed to accurately detect the ICH and IPH subtype in the left image but successfully identified the lesion in the right image. As depicted in Fig.~\ref{fig4}, the performance of three proprietary MLLMs improves as the size of the hemorrhage region increases. This suggests that current MLLMs have limited screening capabilities for small lesions across various ICH subtypes and underscores the disparity in MLLMs' ability to recognize different ICH subtypes. Specifically, they are more proficient in identifying IPH with higher edge contrast and concentrated lesions, whereas their performance declines in recognizing SAH with lower contrast and more dispersed lesions.
Furthermore, when slices contain metal objects or calcifications, MLLMs are prone to misinterpret them as hemorrhages. Additionally, metal artifacts can lead MLLMs to misclassify positive cases as normal due to difficulties in distinguishing visual illusions caused by the artifacts, resulting in misjudgment.
Overall, the predictive accuracy of MLLMs in ICH subtype classification remains significantly lower than that of trained deep classification networks. However, MLLMs can enhance interpretability through language interactions, such as describing hemorrhage locations, offering treatment suggestions, and noting abnormalities. Future fine-tuning of open-source models is expected to improve their performance, potentially contributing positively to triage, examination, monitoring, and prognosis in ICH management \cite{guideline2022, BH4-MR-display}.

\section{Limitation}

This study has two main limitations. Firstly, due to hardware resource constraints, we utilized open-source models with relatively small parameter sizes within their respective series. This may hinder these models from fully leveraging their potential in ICH subtype classification tasks. Secondly, due to the current limitations of MLLMs, our hemorrhage subtype prediction analysis was restricted to 2D slices, potentially causing certain cranial structures to be misidentified as high-density hemorrhage areas. Therefore, advancing high-precision MLLMs capable of volumetric medical image processing is a vital direction for future research. In upcoming work, we aim to develop a structured image-text dataset specifically for ICH subtype classification and investigate the fine-tuning of MLLMs for this task to further improve their performance \cite{yuan2024mixed, yuan2025image, wang2025mamba, yuan2025guided, yuan2024simultaneous, wang2024samihs, yuan2024weighted, yuan2023guided, wang2023coam, yuan2021efficient, yuan2025net2net}.

\section{Conclusions}

This study compares multi-modal large language models (MLLMs) to traditional deep learning methods in intracranial hemorrhage (ICH) subtyping. The findings reveal that although MLLMs excel in certain tasks, their overall accuracy is inferior to deep learning models. Nevertheless, MLLMs enhance interpretability through language interactions, indicating potential in medical imaging analysis. Future efforts will focus on model fine-tuning and developing more precise MLLMs to improve performance in volumetric ICH subtyping.

\section{Code Availability}

The preprocessing data and code used in this research are available on GitHub (\url{https://github.com/mileswyn/ICH_MLLMs_validation}).

\section{Acknowledgments}

This work was supported by the National Natural Science Foundation of China (Grant No. 92148206), the Interdisciplinary Research Support Program of Huazhong University of Science and Technology (Grant No. 2024JCYJ010), the Tongji Hospital “Jiebang Guashuai” Key Task Project (Grant No. 25-2KYC13066-12), and the Hubei Province Central Government-Guided Local Science and Technology Development Special Project (Grant No. 2024BCB111).
%
%

\end{document}